\documentclass[letterpaper, 10 pt, conference]{ieeeconf}  

\IEEEoverridecommandlockouts                              

\usepackage{graphics} 
\usepackage{epsfig} 
\usepackage{mathptmx} 
\usepackage{times} 
\usepackage{amsmath} 
\usepackage{amssymb}  
\usepackage{subcaption}
\usepackage{xcolor}

\title{\LARGE \bf
DS-K3DOM: 3-D Dynamic Occupancy Mapping with Kernel Inference and Dempster-Shafer Evidential Theory
}

\author{
Juyeop Han$^{*1}$, Youngjae Min$^{*2}$, Hyeok-Joo Chae$^{1}$, Byeong-Min Jeong$^{1}$ and Han-Lim Choi$^{1}$   
\thanks{$^{1}$The authors are with the KAIST Institutes for Robotics and the Department of Aerospace Engineering, Korea Advanced Institide of Science and Technology (KAIST), Daejeon, 34141, South Korea {\tt\small \{jyhan, bmjeong, hjchae\}@lics.kaist.ac.kr, hanlimc@kaist.ac.kr}
}%
\thanks{$^{2}$Youngjae Min is with Laboratory for Information and Decision Systems (LIDS), Massachusetts Institute of Technology, Cambridge, MA 02139, USA {\tt\small yjm@mit.edu}}%
\thanks{*equal contributions}
}

\begin{document}

\maketitle
\thispagestyle{empty}
\pagestyle{empty}

\renewcommand{\thefootnote}{\roman{footnote}}
\begin{abstract}
Occupancy mapping has been widely utilized to represent the surroundings for autonomous robots to perform tasks such as navigation and manipulation. While occupancy mapping in 2-D environments has been well-studied, there have been few approaches suitable for 3-D dynamic occupancy mapping which is essential for aerial robots. This paper presents a novel 3-D dynamic occupancy mapping algorithm called DS-K3DOM. We first establish a Bayesian method to sequentially update occupancy maps for a stream of measurements based on the random finite set theory. Then, we approximate it with particles in the Dempster-Shafer domain to enable real-time computation. Moreover, the algorithm applies kernel-based inference with Dirichlet basic belief assignment to enable dense mapping from sparse measurements. The efficacy of the proposed algorithm is demonstrated through simulations and real experiments\footnote{The code is available at: https://github.com/JuyeopHan/dsk3dom\_public}.
\end{abstract}

\section{Introduction}

Understanding the surroundings is of great importance for autonomous robots deployed in unknown or partially known environments to perform tasks such as navigation and manipulation. One of the most central information about the surroundings is occupancy status for the area of interest to recognize target objects or find safe paths to destinations without any collisions. For these purposes, occupancy map is often employed by estimating whether each cell of discretized space or a point over continuous space is occupied by an object or not. When the occupying object is in motion, it is also essential to estimate its dynamic states such as velocity in addition to the occupancy to enable further missions such as collision avoidance and target tracking.

While occupancy mapping in two-dimensional (2-D) environments has been well-studied for ground vehicles, there have been few approaches suitable for 3-D dynamic occupancy mapping which is crucial for aerial robots. Prior to handling dynamic objects, occupancy mapping for static environments already introduces difficulty with sparse and noisy sensor measurements that cause inaccurate occupancy estimation. The problem is more severe in 3-D mapping compared to the 2-D cases as the same number of sensor rays results in sparser coverage in 3-D space. To resolve the issue, various studies have considered spatial correlation among cells, for instance, using Gaussian process regression~\cite{o2012gaussian}, logistic regression with hilbert maps~\cite{doherty2016probabilistic,ramos2016hilbert}, and Bayesian kernel inference~\cite{Doherty2019}. Nevertheless, they are incapable of updating the occupancy map along with the movements of dynamic objects and limited to static environments.

\begin{figure}[tb!]
    \begin{subfigure}{.23\textwidth}
      \centering
      \includegraphics[width=.97\linewidth, height = 1.1in]{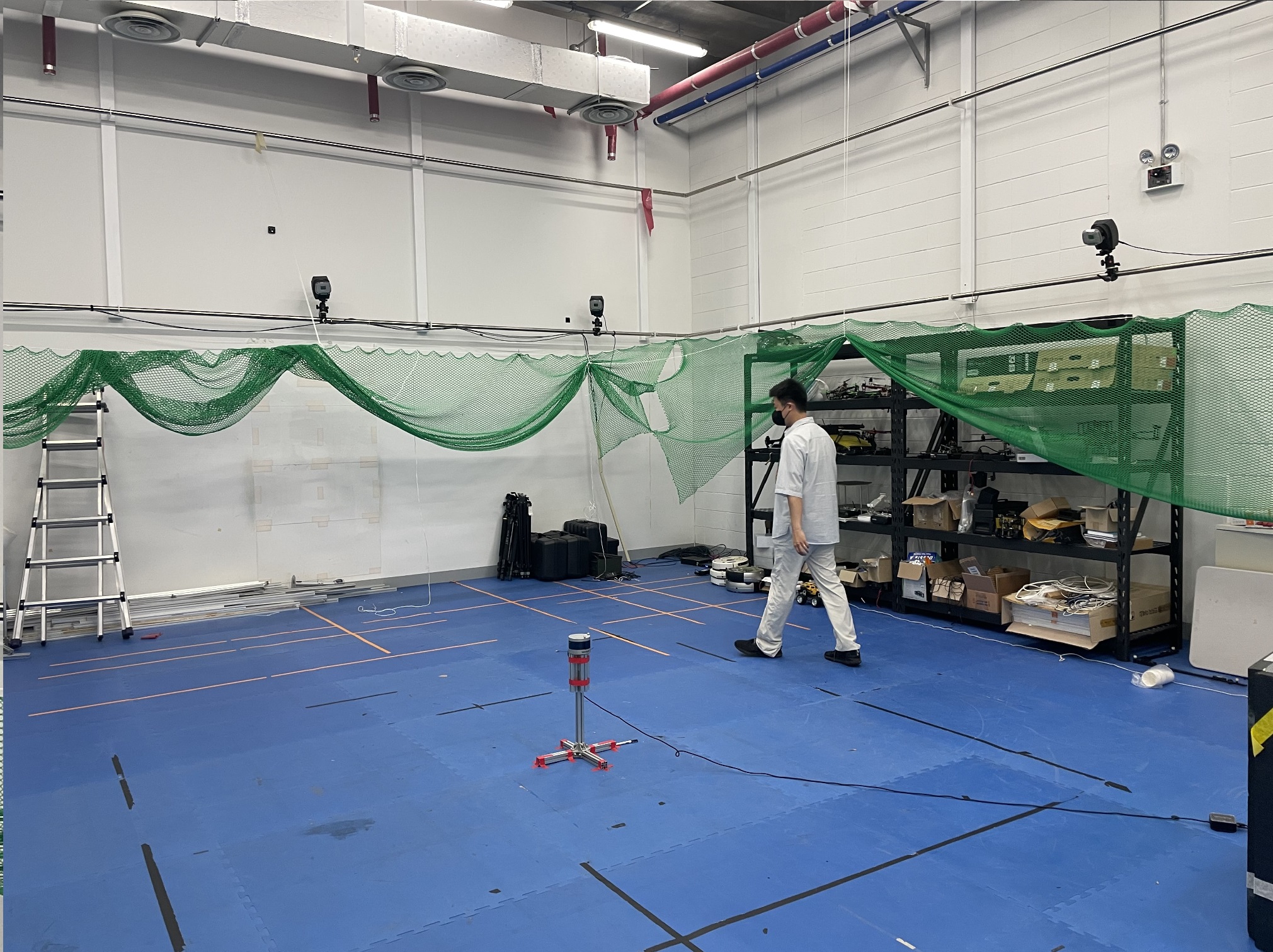}
      \caption{Indoor Environment}
      \label{fig:indoor env}
    \end{subfigure}%
    \hfill
    \begin{subfigure}{.23\textwidth}
      \centering
      \includegraphics[width=.97\linewidth, height = 1.1in]{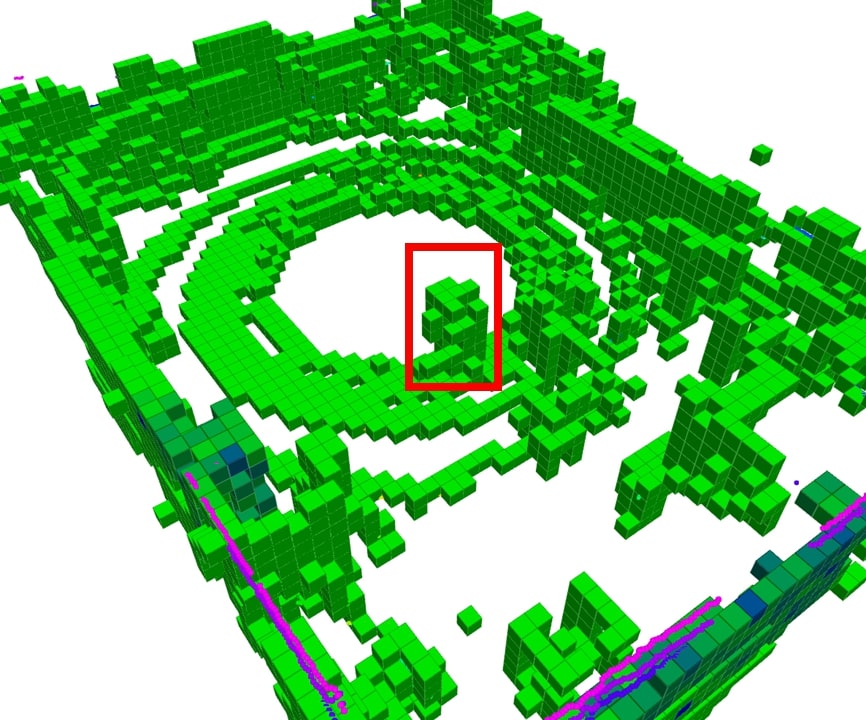}
      \caption{DS-PHD/MIB}
      \label{fig:indoor dsphdmib}
    \end{subfigure}
    \hfill
    \begin{subfigure}{.23\textwidth}
      \centering
      \includegraphics[width=.97\linewidth, height = 1.1in]{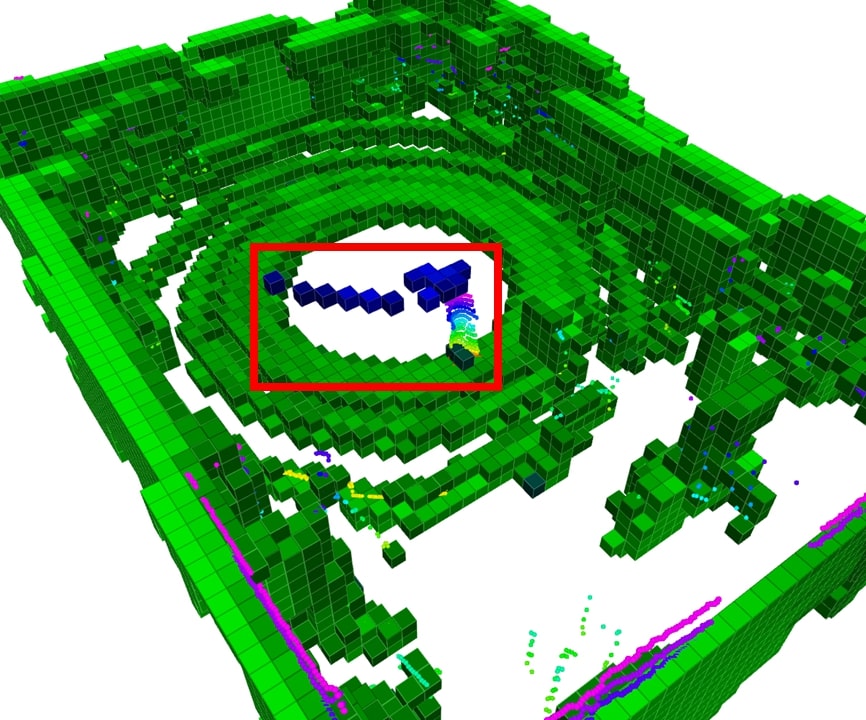}
      \caption{K3DOM}
      \label{fig:indoor k3dom}
    \end{subfigure}
    \hfill
    \begin{subfigure}{.23\textwidth}
      \centering
      \includegraphics[width=.97\linewidth, height = 1.1in]{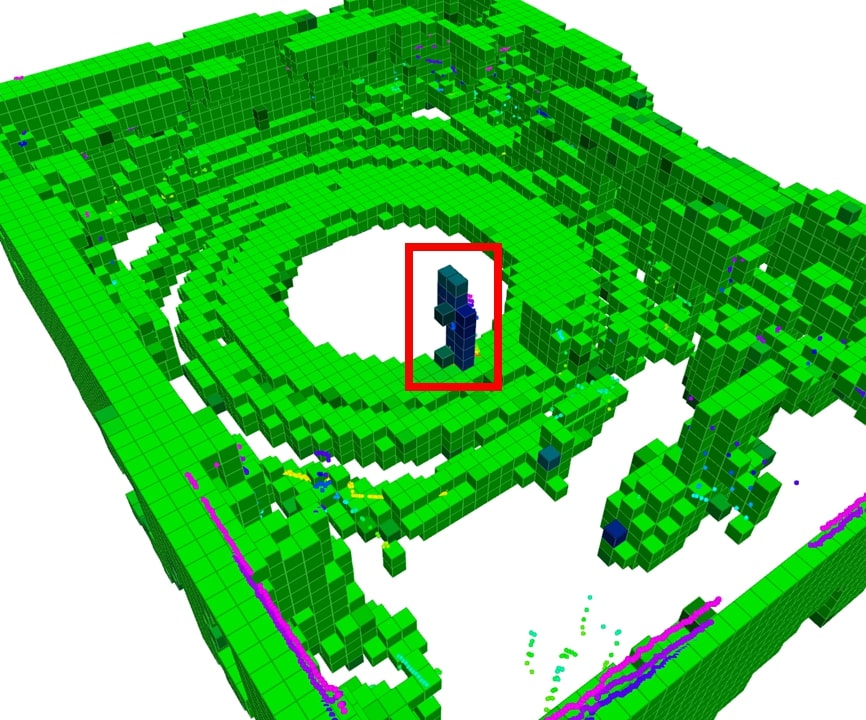}
      \caption{DS-K3DOM}
      \label{fig:indoor dsk3dom}
    \end{subfigure}
\caption{Results of indoor experiments. (a) shows the environment in which a person walking around the LiDAR. (b-d) present the occupancy map estimated by each algorithm. They indicate estimated occupancy with different colors: blue and green for dynamic and static objects, respectively.}
\vspace{-.2in}
\label{fig:Indoor}
\end{figure}

For dynamic environments, several methods have been proposed to accommodate dynamic objects by sequentially updating occupancy maps with a stream of measurements. \cite{o2016gaussian, senanayake2017learning} proposed dynamic occupancy mapping in continuous space using Gaussian process regression, and
\cite{senanayake2017bayesian} took a variational Bayesian approach on Hilbert mapping to learn occupancy maps sequentially. While these methods suffer from high computational cost, \cite{danescu2011modeling} proposed a real-time solution by employing a particle filter. \cite{tanzmeister2016evidential, steyer2018grid} also utilized particle filters in the Dempster-Shafer domain and showed satisfactory performance. Theoretically, \cite{Nuss2018} proposed a rigorous Bayesian method of occupancy mapping, called the PHD/MIB filter, based on the random finite set theory. Also, the authors presented its real-time viable approximation in the Dempster-Shafer domain with particles. However, those methods mainly target 2-D environments, and simply applying them to 3-D environments demands vast memory and computation capabilities due to the higher degree of freedom of 3-D environments. Recently, \cite{Min2021} proposed a real-time solution for 3-D dynamic occupancy mapping, called K3DOM. It efficiently restricts the usage of particles on potential dynamic objects through a kernel-based inference on Dirichlet distribution. Nonetheless, the method is based on heuristics and lacks in rigorous foundation.

Inspired by the mathematical foundation of the PHD/MIB filter and the practicality of K3DOM, this paper presents a novel 3-D dynamic occupancy mapping algorithm called DS-K3DOM. We first theoretically extend the PHD/MIB filter to distinguish dynamic objects from static objects instead of treating them uniformly as occupying objects. Then, we approximate the extended PHD/MIB filter in the Dempster-Shafer domain with particles to enable real-time computation. Taking advantage of the extended structure, we employ particles efficiently to represent potential dynamic objects without unnecessary employment for static objects. In addition, we propose a kernel-based inference with Dirichlet basic belief assignment (BBA)~\cite{Josang2007} to enables dense mapping from the spatially sparse sensor measurements. The efficacy of the proposed algorithm is demonstrated through simulations and real experiments.



\section{Preliminaries}
Before introducing the algorithm, we formulate the problem~(\ref{subsec:problem}) and briefly review the concepts of random finite set (RFS) and probability hypothesis density (PHD)~(\ref{subsec:rfsphd}). For solid background, we refer the readers to~\cite{Nuss2018,Mahler2007,Ristic2016}.

\subsection{Problem Formulation} \label{subsec:problem}

This paper focuses on 3-D dynamic occupancy mapping in discrete space represented with cells. The main problem of the paper is to estimate the occupancy state of each cell among $\Omega := \{D, S, F\}$ at each time-step discretized by step-size $dt\in\mathbb{R}^+$. `$D$' (resp. `$S$') represents the state occupied by a dynamic (resp. static) object, while `$F$' represents the unoccupied (free) state. The ingredients for the state estimation at time-step $k$ are accumulated range sensor measurements $Z_{1:k}:=\{X_t, Y_t\}^k _{t=1}$. $X_t$ is a set of measured locations in $\mathbb{R}^3$, and $Y_t$ is a set of corresponding measured values in $\{0,1\}$ at each time-step $t$. The values of '0' and '1' denote free and occupied measurement, respectively. Note that range sensors indirectly measure free space information. Thus, the closest point on each measurement ray from the query point is utilized as a free measurement as in~\cite{Doherty2019}.

\subsection{Random Finite Set and Probability Hypothesis Density} \label{subsec:rfsphd}
An RFS $X = \{x_1,..., x_n\}$ is a random variable having a value as a finite set. The set consists of random vectors $x_i \in \mathbb{X}$ with a random cardinality $n$ which follows the distribution $\rho(n) := P(\vert X \vert=n)$. For each $n > 0$, let $f_n(x_1,...,x_n)$ denote the symmetric joint probability distribution function (PDF) of the elements of the RFS $X$. Then, the PDF of $X$, $\pi(X)$, and its integration are defined as below. 
\begin{align}
\label{RFS pdf}
 \pi(X = \{x_1,...,x_n\}) = n! \cdot \rho(n) \cdot f_n(x_1,...,x_n),\; \\
 \int \pi(X)\delta X = \pi(\emptyset) + \sum_{n=1}^{\infty}\frac{1}{n!}\int\pi(\{x_1,...,x_n\})dx_1 \cdots dx_n.
\end{align}
Note that $\pi(X = \emptyset) = \rho(0)$. Then, the PHD of the RFS $X$, $D(x)$, is defined over the state-space $\mathbb{X}$ as
\begin{equation}
\label{PHD}
     D(x) = E[\delta_X(x)] = \int \delta_X(x)\pi(X)\delta X,
\end{equation}
where $\delta_X(x) := \sum_{s \in X}\delta_s(x)$ with a Dirac delta function $\delta_s(x)$ concentrated at each $s\in X$. The integration of $D(x)$ over an area results in the expected number of objects in the area.

\section{Extended PHD/MIB Filter}
\label{Sec:2}
In this section, we present the theoretical foundation of DS-K3DOM by extending the PHD/MIB filter \cite{Nuss2018}. The PHD/MIB filter estimates a dynamic occupancy map via Bernoulli RFSs and their joint PHD. It represents the occupancy status of each cell with a Bernoulli RFS and dynamically updates the occupancy estimation by repeating prediction and update steps. For the prediction step, the filter converts the Bernoulli RFSs of all cells into their joint PHD and predicts its change due to the movements of dynamic objects. Then, the joint PHD is approximated back to separate Bernoulli RFSs to update the estimation based on the measurement information on each cell.
We extend this PHD/MIB filter to subdivide the feasible occupancy states of each cell into $\Omega = \{D, S, F\}$. Note that the original filter does not distinguish between the dynamic and static states and treats them uniformly as `occupied'.

\subsection{Representation of Objects via Bernoulli RFS and PHD} \label{subsec:representation}

We represent the multi-object distribution of the surrounding objects using Bernoulli RFSs, where an element $x=[p^T \; v^T]^T\in \mathbb{X}$ of an RFS represents the state of a point object with its position $p \in \mathbb{R}^3$ and velocity $v \in \mathbb{R}^3$. Under the assumption of having at most one object in each cell, the Bernoulli RFS of each cell ($c$) is modeled to have PDF as 
\begin{equation}
\label{PDF per cell}
    \pi^{(c)}(X) = \begin{cases}
        1 - r^{(c)} & \text{if $X = \emptyset$} \\
        r^{(c)} \cdot p^{(c)}(x)  & \text{if $X = \{x\}$}\\
        0 & \text{if $\vert X \vert \geq 2 $}
    \end{cases},
\end{equation}
where $r^{(c)}$ and $p^{(c)}(x)$ are the existence probability and PDF of an object, respectively. Regarding the mutually exclusive events of the object being static and dynamic, we further decompose the probabilities and PDFs as
\begin{equation}
\begin{aligned}
    r^{(c)} &= r_D^{(c)} + r_S^{(c)},\\
    p^{(c)}(x) &= (r_D^{(c)} \cdot p_D^{(c)}(x)+r_S^{(c)} \cdot p_S^{(c)}(x))/r^{(c)},
\end{aligned}
\end{equation}
where $r_D^{(c)}$ (resp. $r_S^{(c)}$) is the existence probability of the dynamic (resp. static) portion, and $p_D^{(c)}$ (resp. $p_S^{(c)}$) is its PDF. Then,
the PHD of the Bernoulli RFS is also decomposed as
\begin{equation}
    D^{(c)}(x) = r^{(c)} \cdot p^{(c)}(x) = D_D^{(c)}(x) + D_S^{(c)}(x),
\end{equation}
if we consider separate Bernoulli RFSs for the dynamic and static portions with their PHDs $D_D^{(c)}(x) = r_D^{(c)} \cdot p_D^{(c)}(x)$ and $D_S^{(c)}(x) = r_S^{(c)} \cdot p_S^{(c)}(x)$, respectively. Their joint PHD is
\begin{equation}\label{eq:joint_PHD}
    D(x) = \sum_{c} D^{(c)}(x)
    = D_D(x) + D_S(x),
\end{equation}
where $D_D(x):=\sum_{c} D_D^{(c)}(x)$ and $D_S(x):=\sum_{c} D_S^{(c)}(x)$.

\subsection{Bayesian Update of Bernoulli RFS}\label{subsec:extended_phdmib}

The extended PHD/MIB filter dynamically updates the Bernoulli RFSs according to the new measurements in a Bayesian framework. It first predicts the changes in the joint PHD of the Bernoulli RFSs of all cells by propagating the current estimation. Then, it approximately converts the predicted PHD back to the new Bernoulli RFSs and updates their posterior distributions with the measurement information. Unlike the PHD/MIB filter, our extended filter utilizes the decomposed structures for the dynamic and static portions that enables efficient particle realization in Section~\ref{sec:dsk3dom}.

In the prediction step, the movements of dynamic objects are the main source of change. Thus, we follow the standard prediction step of a PHD filter~\cite{Mahler2003} for the dynamic object, while the static portion remains the same. We represent the predicted joint PHD at time-step $k+1$ in a separated structure $D_+(x_{k+1})=D_{D+}(x_{k+1}) + D_{S+}(x_{k+1})$ as in~\eqref{eq:joint_PHD}. Then, each portion is predicted as:
\begin{align}
    D_{D+}(x_{k+1}) &= D_{b,+}(x_{k+1})+p_S \int f_{D+}(x_{k+1} \vert x_{k})D_{D}(x_k)dx_k,\label{eq:phd_pred_dyn} \\
    D_{S+}(x_{k+1}) &= D_{S}(x_{k+1}),\label{eq:phd_pred_stat}
\end{align}
with the persistence probability $p_S$ and the transition density of a dynamic object $f_{D+}(x_{k+1} \vert x_{k})$. $D_{b,+}(x_{k+1})$ is a newly generated PHD through the birth process that could reduce false negative occupancy estimation, which is usually more unfavorable than false positive one in many applications such as collision avoidance. We denote the last term in~\eqref{eq:phd_pred_dyn} as $D_{p,+}(x_{k+1})$ to indicate the predicted PHD for persisting dynamic object.

Next, we derive new Bernoulli RFSs that corresponds to the predicted PHD. From the definition of PHD, the existence probabilities of the persistent dynamic object and static object in each cell ($c$) are, respectively,
\begin{align}
    r_{p,+}^{(c)} &= \min(\int_{x_{k+1} \in c}D_{p,+}(x_{k+1})dx_{k+1}, 1),\label{eq:dyn_pred_prob} \\
    r_{S+}^{(c)} &= \min(r_S^{(c)}, 1 - r_{p,+}^{(c)}),\label{eq:stat_pred_prob}
\end{align}
where $x_{k+1}\in c$ indicates that the position of $x_{k+1}$ is inside the cell $(c)$.
The probabilities are clipped so that each of them and their sum do not exceed 1. Meanwhile, the existence probability of new-born dynamic objects is modeled as
\begin{equation}
\label{prediction probs.}
    r_{b,+}^{(c)} = p_B \cdot (1 - r_{p,+}^{(c)}-r_{S+}^{(c)}),
\end{equation}
given the prior birth probability $p_B\in[0,1]$. Then, we compute the PDF of the predicted objects inside the cell $(c)$ as
\begin{equation}
    p_{(\cdot)}^{(c)}(x_{k+1}) = D_{(\cdot)}(x_{k+1})/r_{(\cdot)}^{(c)}
\label{predicted PDF}
\end{equation}
for nonzero $r_{(\cdot)}^{(c)}$ where $(\cdot)$ corresponds to `$b,+$', `$p,+$', and `$S+$' for new-born dynamic, persistent dynamic, and static object, respectively. With these parameters, the Bernoulli RFS of the object in each cell $(c)$ is predicted to have PDF
\begin{equation*}
\begin{split}
    &\pi_{+}^{(c)}(X_{k+1}) = \\
    &\begin{cases}
       1 - r_{p,+}^{(c)} - r_{b,+}^{(c)} - r_{S+}^{(c)} & \text{if $X_{k+1} = \emptyset$} \\
        \begin{array}{c}
        r_{b,+}^{(c)}  p_{b,+}^{(c)}(x_{k+1}) + r_{p,+}^{(c)} p_{p,+}^{(c)}(x_{k+1})\\
        + r_{S+}^{(c)} p_{S+}^{(c)}(x_{k+1})
        \end{array}
        & \text{if $X_{k+1} = \{x_{k+1}\}$}\\
        0 & \text{if $\vert X_{k+1} \vert \geq 2 $}
    \end{cases}
\end{split}.
\end{equation*}

In the update step, the predicted PDF and new measurements $Z_{k+1}$ are utilized to update the posterior PDF of Bernoulli RFS via the Bayes' rule~\cite{Ristic2016}:
\begin{equation}
\label{RFS bayes rule}
    \pi^{(c)}(X_{k+1} \vert Z_{k+1})
    = \frac{\eta(Z_{k+1} \vert X_{k+1})\pi^{(c)}_{+}(X_{k+1})}{\int \eta(Z_{k+1} \vert X_{k+1})\pi^{(c)}_{+}(X_{k+1}) \delta X_{k+1}},
\end{equation}
where $\eta(Z_{k+1} \vert X_{k+1})$ denotes the likelihood function for the measurements $Z_{k+1}$ given the posterior RFS $X_{k+1}$. Omitting the conditional part for notational simplicity, the joint posterior PHD is given by the sum of all Bernoulli RFS instances:
\begin{equation}
    D_{k+1}(x_{k+1}) = \sum_{c}{\pi^{(c)}(X_{k+1} = \{x_{k+1}\})}.
\end{equation}

\section{DS-K3DOM: Approximation of Extended PHD/MIB filter in Dempster-Shafer Domain}\label{sec:dsk3dom}

\begin{figure}
      \centering
      \includegraphics[width=0.95\linewidth]{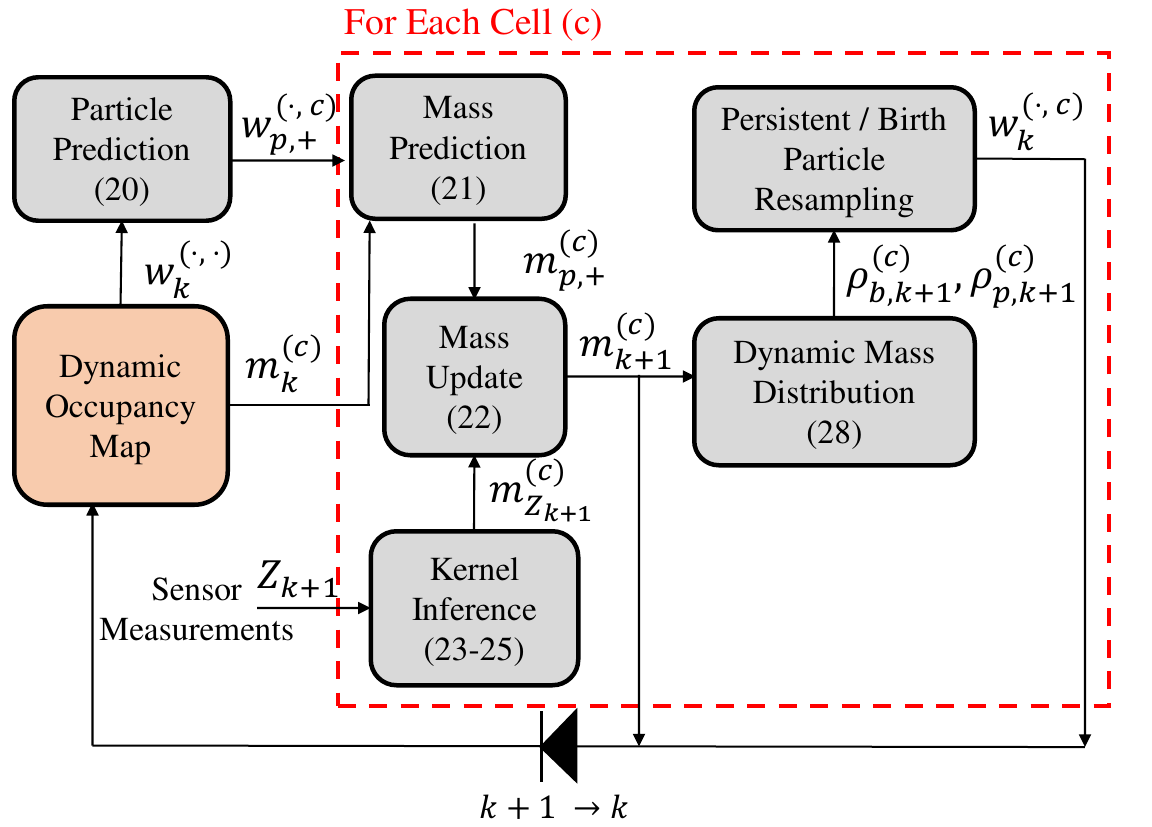}
      \caption{Framework of DS-K3DOM}
  \label{fig:DSK3DOM Framework}
  \vspace{-.2in}
\end{figure}

In this section, we propose DS-K3DOM shown in Fig.~\ref{fig:DSK3DOM Framework} which is an efficient particle realization of the extended PHD/MIB filter in the Dempster-Shafer (DS) domain similar to~\cite{Nuss2018}. It approximates the practically intractable computations in Section~\ref{subsec:extended_phdmib} and enables real-time operations of the filter. By doing so, we economically employ particles only for potential dynamic objects in the observed area without their unnecessary usages for the unobserved area or static objects. Moreover, we exploit the multi-hypothesis structure of the Dempster-Shafer evidential theory (DST) to deal with the ambiguous sensor observations which do not distinguish between dynamic and static statuses. 
In addition, we adopt Dirichlet BBA \cite{Josang2007} in the DS domain to utilize a kernel-based inference, as in~\cite{Min2021}, to enable dense mapping from the sparse sensor measurements. The omitted information on the particle management such as its resampling process follows~\cite{Nuss2018}. 

\subsection{Dempster-Shafer Evidential Theory}
We first briefly review the concept of the DST; for detailed knowledge, we refer the readers to~\cite{Rakowskey2007}. Let a hypothesis be a feasible and exclusive incident in a situation and a frame of discernment be the finite set $\Omega$ that consists of all possible hypotheses. A mapping called a basic belief assignment (BBA) or mass $m:2^{\Omega} \rightarrow [0,1]$ represents the normalized degree of evidence for a set of hypotheses, satisfying the following conditions:
\begin{equation}
\label{Dempster-Shafer}
    m(\emptyset) = 0 \text{ and} \sum_{X \subseteq \Omega}m(X) = 1.
\end{equation}
A set $X$ that satisfies $m(X) > 0$ is called a focal element.
Then, DST provides a rule to update the evidential belief. Dempster's rule of combination is an operation that combines a pair of independent BBAs from multiple data sources that share the same frame of discernment $\Omega$. The rule of combination is represented by
\begin{equation}
\label{DS rule of combination}
    (m_1 \oplus m_2)(X) = \frac{\sum_{A \cap B = X} m(A)m(B)}{1 - \sum_{A \cap B = \emptyset} m(A)m(B)}
\end{equation}
with $A, B, X \subseteq \Omega$.

\subsection{Representation of Cell States in Dempster-Shafer Domain}

We consider observations from a range sensor that only provide evidence on whether a cell is either occupied or free. Accordingly, the BBA $m(\{D, S\})$ (resp. $m(\{F\})$) indicates the evidences supporting that the cell is occupied (resp. free), while $m(\Omega)$ indicates the degree of uncertainty over the cell state. Thus, we define the set of all focal elements as $\Xi := \{ \{D\}, \{S\}, \{F\}, \{D, S\}, \Omega \}$ and compute the probability of each occupancy state for each cell $(c)$ via the pignistic probabilities~\cite{Smets2000}:
\begin{equation}
\label{pignistic transformation}
    P^{(c)}(X) = \sum_{Y \in \Xi}{\frac{\vert X \cap Y \vert}{\vert Y \vert}m(Y)} \,\,\,\, \forall X \in \Omega
\vspace{-.05in}
\end{equation}

Unlike \cite{Nuss2018}, DS-K3DOM employs particles only for dynamic objects so that their weights in each cell $(c)$ add up to the dynamic mass $m_k^{(c)}(\{D\})$. The PDF of a dynamic point object $p_{D,k}^{(c)}(x_k)$ is then approximated as below:
\begin{equation} \label{Dynamic cell particle distribution}
\begin{gathered}
    m_k^{(c)}(\{D\}) = \sum_{i=1}^{\nu_k^{(c)}}w_k^{(i,c)}, \\
    p_{D,k}^{(c)}(x_k) \approx \frac{1}{m_k^{(c)}(\{D\})} \sum_{i=1}^{\nu_k^{(c)}}w_k^{(i,c)} \delta(x_k - x_k^{(i,c)}),
\end{gathered}
\vspace{-.1in}
\end{equation}
where $\nu_k^{(c)}$ indicates the number of particles of each cell $(c)$ and $w_k^{(i,c)}$ and $x_k^{(i,c)}$ are the weight and state of the $i$-th particle in the cell, respectively.
 
\subsection{Prediction of Particles and Masses}
We adopt a constant velocity (CV) model to propagate each particle in the prediction step. The state of each persistent particle at the time-step $k+1$ is predicted to evolve as 
\begin{equation}
\label{CV model}
    x_{p,+} = 
    \left[\begin{matrix}
    I_3 & dt \cdot I_3 \\
    \mathbf{0}_{3 \times 3} & I_3
    \end{matrix}\right]x_k
    + n_x
\end{equation}
with a process noise $n_x = [n_p^T \ n_v^T]^T$, $n_p \sim N(\mathbf{0}_3, \sigma_p I_3)$, and $n_v \sim N(\mathbf{0}_3, \sigma_v I_3)$. As described in~\ref{subsec:representation}, the first three dimensions are for position while the other three are for velocity. The propagated particles are then newly indexed so that each cell $(c)$ contains $\nu_{p,+}^{(c)}$ particles.

The prediction of each focal element's mass is designed to imitate the prediction step in the extended PHD/MIB filter,  (\ref{eq:dyn_pred_prob}-\ref{eq:stat_pred_prob}), and to properly distribute $m_k^{(c)}(\{D,S\})$ to $m_{p,+}^{(c)}(\{D\})$ and $m_{p,+}^{(c)}(\{S\})$ as below:
\begin{equation}
\label{mass prediction}
\begin{split}
    m^{(c)}_{p,+}(\{D\}) = &\min \Big( 1, \hspace{0.1cm}\sum_{i=1}^{\nu_{p,+}^{(c)}}w_{p,+}^{(i,c)}
    + \delta^{(c)}_k\beta\gamma^{dt}m_k^{(c)}(\{D,S\}) \Big), \\
    m^{(c)}_{p,+}(\{S\}) = &\min \Big( 1-m_{p,+}^{(c)}(\{D\}), \\
    & \gamma^{dt} m^{(c)}_k(\{S\}) +(1-\delta^{(c)}_k)\beta\gamma^{dt}m_k^{(c)}(\{D,S\}) \Big), \\
    m^{(c)}_{p,+}(\{D,S\}) = &\min \Big(1 - m^{(c)}_{p,+}(\{D\}) - m^{(c)}_{p,+}(\{S\}), \\ 
    &\hspace{2.7cm}(1-\beta)\gamma^{dt}m_k^{(c)}(\{D,S\}) \Big),\\
    m^{(c)}_{p,+}(\{F\}) = &\min \Big( 1 - m^{(c)}_{p,+}(\{D\})-m^{(c)}_{p,+}(\{S\})\\
    &\hspace{1.6cm} - m^{(c)}_{p,+}(\{D,S\}),\hspace{0.1cm} \gamma^{dt}m_k^{(c)}(\{F\}) \Big),\\
    m^{(c)}_{p,+}(\Omega) = &1 - \sum_{X \in 2^\Omega/\Omega}m^{(c)}_{p,+}(X).
\end{split}
\end{equation}
$\gamma$ denotes the decaying factor on the beliefs due to the increased uncertainty as time passed. We distribute the $\beta\in[0,1]$ portion of the decayed occupied mass $\gamma^{dt} m_k^{(c)}(\{D,S\})$ to the dynamic mass $m_{p,+}^{(c)}(\{D\})$ and the static mass $m_{p,+}^{(c)}(\{S\})$. Specifically, the $\delta_k^{(c)}$ portion of the mass $\beta\gamma^{dt} m_k^{(c)}(\{D,S\})$ is assigned to $m_{p,+}^{(c)}(\{D\})$, and the rest is assigned to $m_{p,+}^{(c)}(\{S\})$. Since the mass of the occupied hypothesis $m_k^{(c)}(\{D,S\})$ is a potential for the dynamic and static states, we determine the spliting ratio $\delta_k^{(c)}:= P^{(c)}_k(D)/(P^{(c)}_k(D) + P^{(c)}_k(S))$ in proportion to the probability of each state. The redistribution procedure is critical to spread out the occupied mass $m_k^{(c)}(\{D, S\})$ accumulated from the sensor observations since all particles are assigned only for the dynamic mass $m_k^{(c)}(\{D\})$. Finally, the clipping on the masses guarantees the normalization condition~(\ref{Dempster-Shafer}).

\subsection{Measurement Update using Dirichlet BBA with Kernel Inference}
Given a set of sensor measurements $Z_{k+1} = \{X_{k+1}, Y_{k+1}\}$, we build an observation BBA $m_{Z_{k+1}}^{(c)}$ and combine it with the predicted BBA to update the posterior BBA through~\eqref{DS rule of combination}:
\begin{equation}
\label{information fusion}
    m_{k+1}^{(c)} = m_{p,+}^{(c)} \oplus m_{Z_{k+1}}^{(c)}.
\end{equation}
However, the sparse range sensor measurements in the 3-D environment make it hard to generate dense estimation. We overcome this challenge by inferring the spatial information around the sensor observations using a kernel method similarly to~\cite{Min2021}. While the method in~\cite{Min2021} is based on the Bayes' rule and cannot be applied in the DS domain, we propose a new kernel inference using Dirichlet BBA \cite{Josang2007} to generate a dense observation BBA $m_{Z_{k+1}}^{(c)}$. 

Let $Dir(\vec{\alpha}^{(c)})$ be a Dirichlet distribution of all focal elements of $\Omega$, where $\vec{\alpha}^{(c)} = [\alpha^{(c)}(x_1), \cdots,  \alpha^{(c)}(x_n)]^T$ is the parameter vector for all focal elements $x_1, \cdots, x_n \in \Xi$. Each element $\alpha^{(c)}(x_i)$ is the sum of a prior evidence $r_0(x_i)$ and an observation evidence $r^{(c)}(x_i)$ given the sum of all prior evidences $R_0$:
\begin{equation}
\label{Dirichlet parameter}
    \alpha^{(c)}(x_i) = r_0(x_i) + r^{(c)}(x_i) \text{ and } R_0 = \sum_{X_i \in \Xi}r_0(X_i).
\end{equation}
We compute the observation evidences from the sensor measurements $Z_{k+1}$ using a kernel inference. For a kernel function $k:\mathbb{R}^3\times\mathbb{R}^3\rightarrow\mathbb{R}$ and the center of the cell $x^{(c)}$,
\begin{equation}
\label{kernel evidence}
\begin{split}
    r^{(c)}(\{D,S\}) &= \sum_{(x,y) \in Z_{k+1}}k(x^{(c)},x)y, \\
    r^{(c)}(\{F\}) &= \sum_{(x,y) \in Z_{k+1}}k(x^{(c)},x)(1-y).
\end{split}
\end{equation}
Other than the occupied and free focal elements, i.e., if $x_i\neq\{D, S\}$ and $x_i\neq\{F\}$, $r^{(c)}(x_i) = 0$. We use the kernel function employed in~\cite{Doherty2019} and \cite{Min2021}:
\begin{equation*}
\begin{split}
    &k(x,x') := \\
    &\begin{cases}
        \sigma_0[\frac{1}{3}(2+\cos{(2\pi\frac{d}{l}}))(1-\frac{d}{l})+\frac{1}{2\pi}\sin{(2\pi\frac{d}{l})}] & \text{if $d<l$} \\
        0 & \text{if $d \geq l$}
    \end{cases},
\end{split}
\end{equation*}
with $d := \Vert x - x' \Vert_2$, a kernel scale $\sigma_0$, and a length scale $l$. Then, the observation Dirichlet BBA is computed based on the parameters to satisfy the conditions~\eqref{Dempster-Shafer}:
\begin{equation}
\label{Dirichlet BBA}
m^{(c)}_{Z_{k+1}}(x_i) = \begin{cases}
    r^{(c)}(x_i)/\sum_{x_j \in \Xi}\alpha^{(c)}(x_j) & \text{if $x_i \in \Xi/\{\Omega$\}} \\
    R_0/\sum_{x_j \in \Xi}\alpha^{(c)}(x_j) & \text{if $x_i = \Omega$}
\end{cases}.
\end{equation}

\subsection{Update of BBA and mass of the Birth Dynamic Object}
While the updated dynamic mass $m_{k+1}^{(c)}(\{D\})$ is redistributed to the weights of the persistent particles in the cell, we use some portion of it for the birth process of new particles as in the extended PHD/MIB filter. As in~\eqref{eq:phd_pred_dyn}, the
dynamic mass $m_{k+1}^{(c)}(\{D\})$ is decomposed to the persistent dynamic mass $\rho^{(c)}_{p,k+1}$ and the birth dynamic mass $\rho^{(c)}_{b,k+1}$, i.e. $m_{k+1}^{(c)}(\{D\}) = \rho^{(c)}_{p,k+1} + \rho^{(c)}_{b,k+1}$.

We determine the ratio between the two masses according to the extended PHD/MIB filter. Assume that the ratio between the posterior probabilities of a persistent and a birth dynamic object, $r^{(c)}_{p,k+1}$ and $r^{(c)}_{b,k+1}$, respectively, in the extended PHD/MIB filter are the same as that of their corresponding predicted probabilities, $r^{(c)}_{p,+}$ and $r^{(c)}_{b,+}$ as in~\cite{Nuss2018}. Then from~\eqref{prediction probs.}, the ratio is
\begin{equation}
\label{pers-birth ratio}
\frac{r^{(c)}_{b,k+1}}{r^{(c)}_{p,k+1}}
= \frac{r^{(c)}_{b,+}}{r^{(c)}_{p,+}}
= \frac{p_B(1-r^{(c)}_{p,+}-r^{(c)}_{S+})}{r^{(c)}_{p,+}}.
\end{equation}
Following this ratio~\eqref{pers-birth ratio}, we set the ratio between $\rho^{(c)}_{p,k+1}$ and $\rho^{(c)}_{b,k+1}$ similarly as
\begin{equation}
\label{pers-birth mass ratio}
\frac{\rho^{(c)}_{b,k+1}}{\rho^{(c)}_{p,k+1}}
= \frac{p_B(1-m^{(c)}_{p,+}(\{D\})-m^{(c)}_{p,+}(\{S\}))}{m^{(c)}_{p,+}(\{D\})}.
\end{equation}
Therefore, the posterior dynamic masses are derived as
\begin{equation}
\label{newborn and persistent paticle}
\begin{split}
    \rho^{(c)}_{b,k+1}
    &= \frac{m_{k+1}^{(c)}(\{D\}) \cdot p_B(1-m^{(c)}_{p,+}(\{D\})-m^{(c)}_{p,+}(\{S\}))}
    {p_B(1-m^{(c)}_{p,+}(\{D\})-m^{(c)}_{p,+}(\{S\})) + m^{(c)}_{p,+}(\{D\})},\\
    \rho^{(c)}_{p,k+1} &= m_{k+1}^{(c)}(\{D\}) - \rho^{(c)}_{b,k+1}.
\end{split}
\end{equation}
\vspace{-.15in}

\begin{figure*}[t]
    \hfill
    \begin{subfigure}{.15\textwidth}
      \centering
      \includegraphics[width=\linewidth]{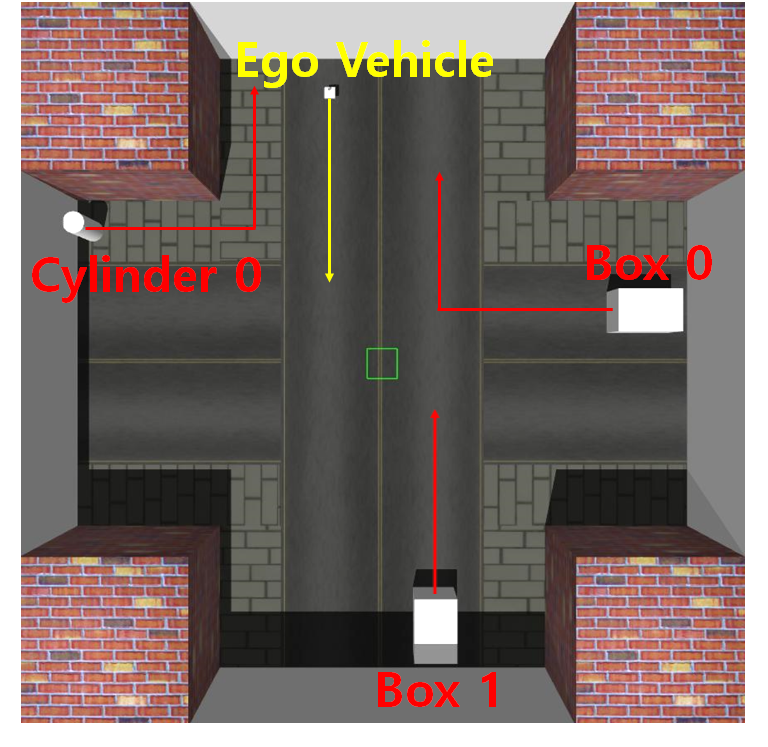}
      \caption{Environment}
      \label{fig:simulation environment}
    \end{subfigure}
    \hfill
    \begin{subfigure}{.205\textwidth}
      \centering
      \includegraphics[width=\linewidth]{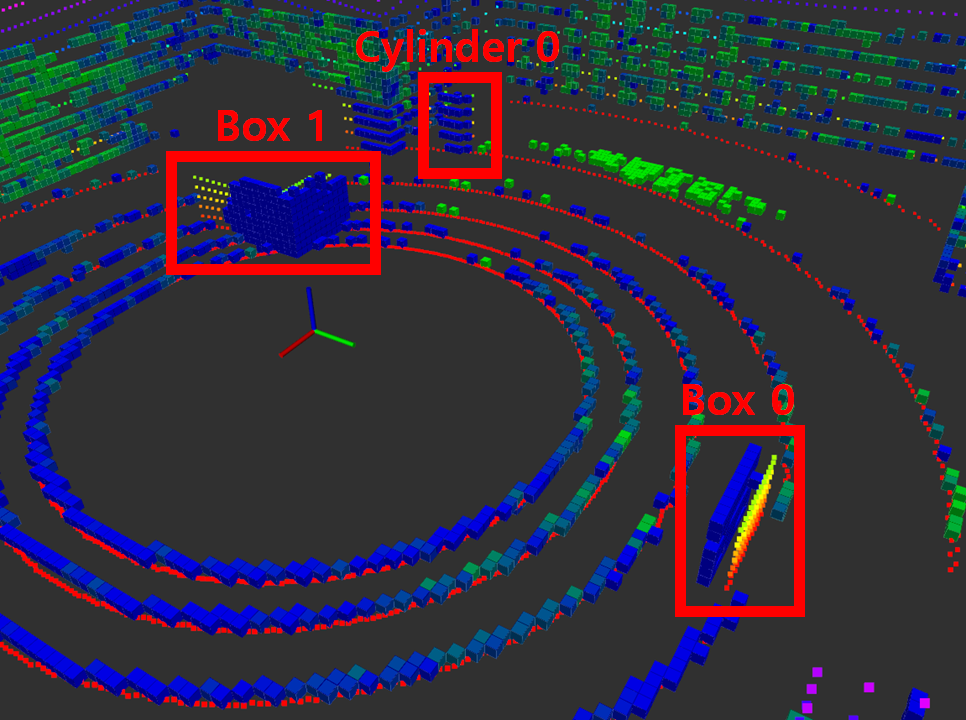}
      \caption{DS-PHD/MIB}
      \label{fig:simulation dsphdmib}
    \end{subfigure}
    \hfill
    \begin{subfigure}{.205\textwidth}
      \centering
      \includegraphics[width=\linewidth]{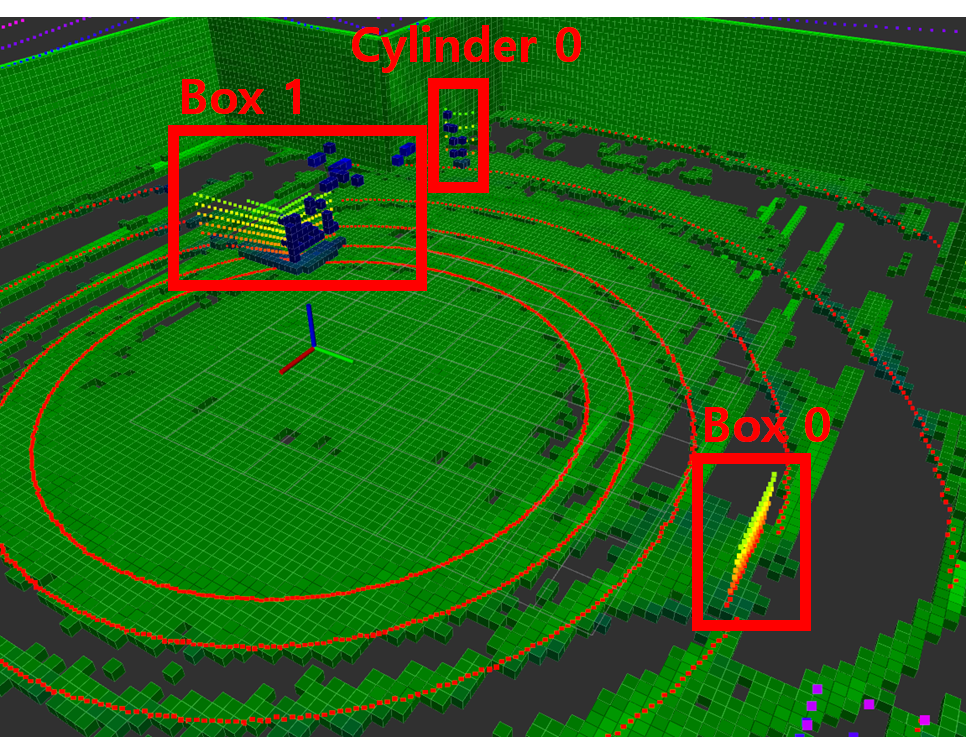}
      \caption{K3DOM}
      \label{fig:simulation k3dom}
    \end{subfigure}
    \hfill
    \begin{subfigure}{.205\textwidth}
      \centering
      \includegraphics[width=\linewidth]{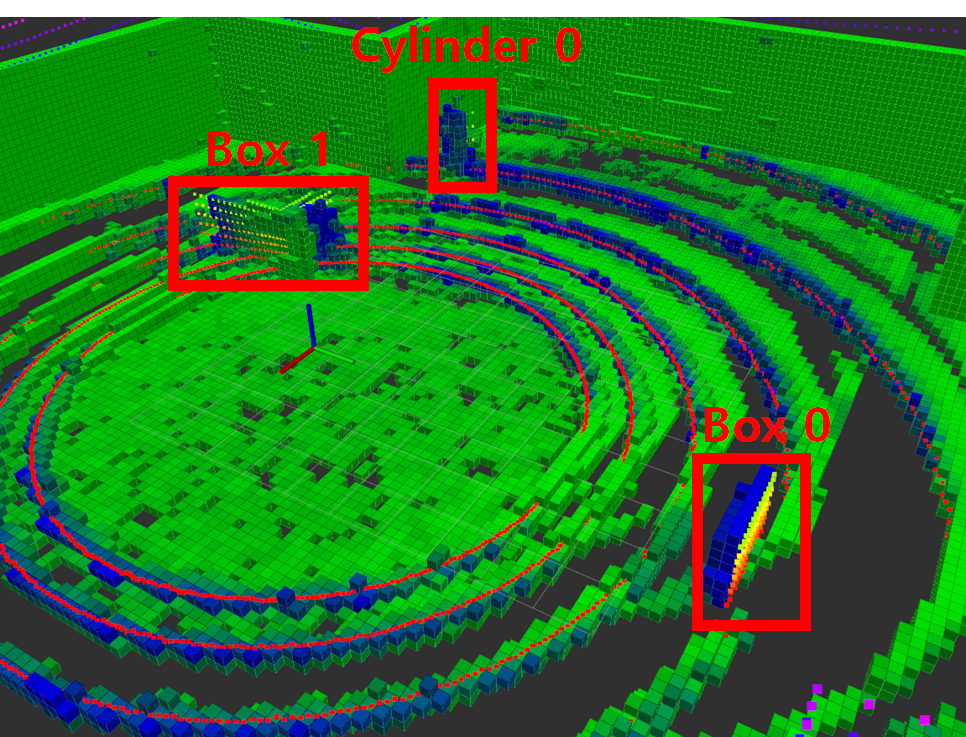}
      \caption{DS-K3DOM}
      \label{fig:simulation naive_ds_k3dom}
    \end{subfigure}
    \hfill
    \begin{subfigure}{.205\textwidth}
      \centering
      \includegraphics[width=\linewidth]{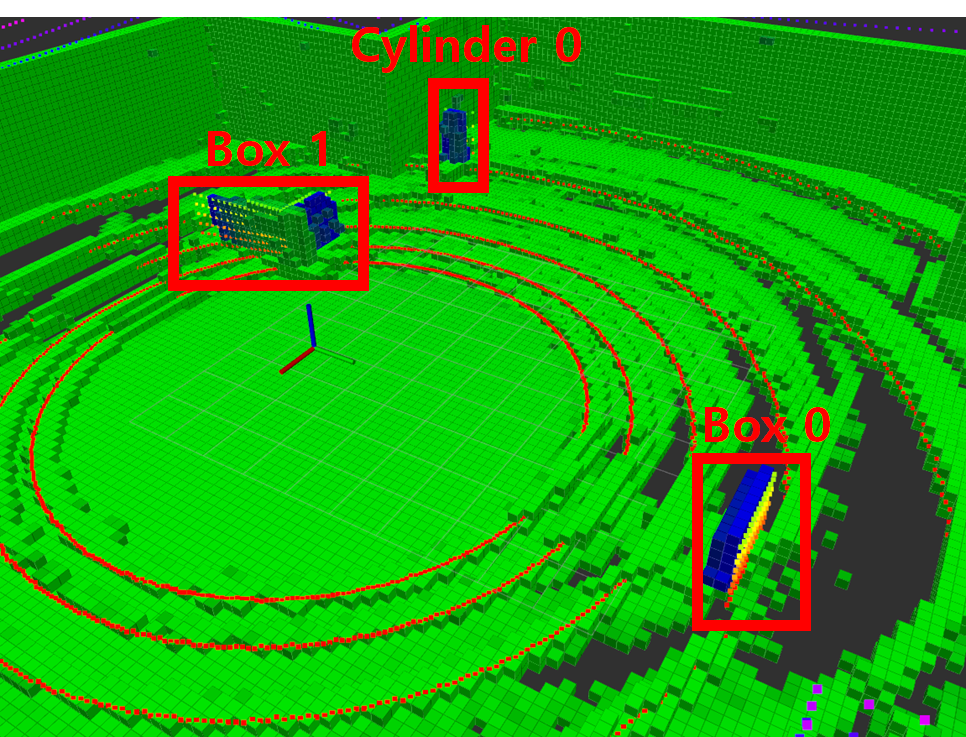}
      \caption{DS-K3DOM (F)}
      \label{fig:simulation ds_k3dom}
    \end{subfigure}
    \label{fig:simulation and mapping}
    \hfill
    \begin{subfigure}{.19\textwidth}
      \centering
      \includegraphics[width=\linewidth]{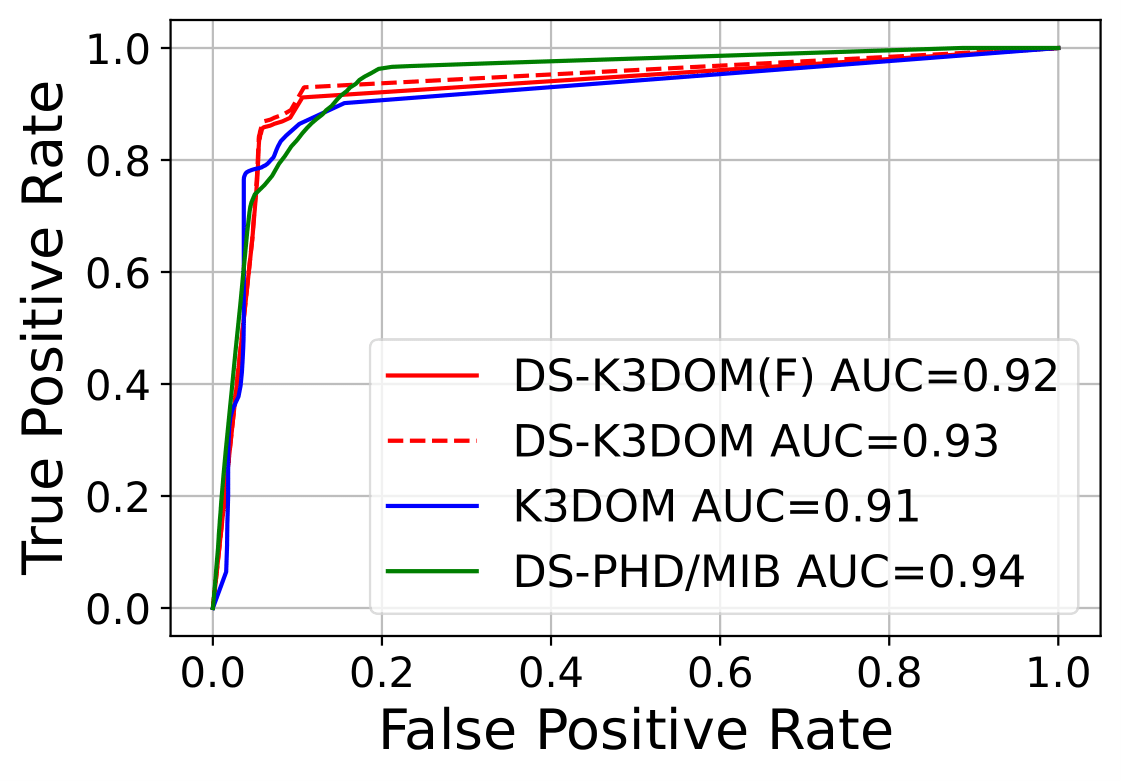}
      \caption{Occupancy ROC curve}
      \label{fig:occupy ROC}
    \end{subfigure}
    \hfill
    \begin{subfigure}{.19\textwidth}
      \centering
      \includegraphics[width=\linewidth]{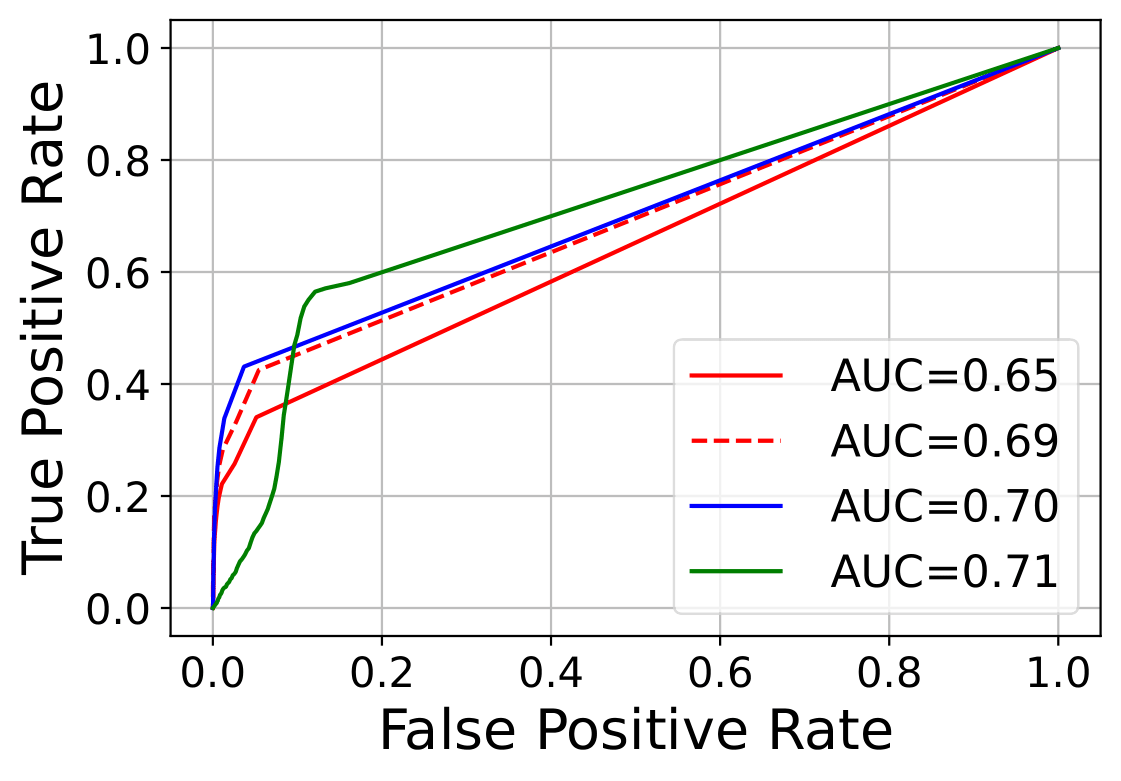}
      \caption{Dynamic ROC curve}
      \label{fig:dynamic ROC}
    \end{subfigure}
    \hfill
    \begin{subfigure}{.19\textwidth}
      \centering
      \includegraphics[width=\linewidth]{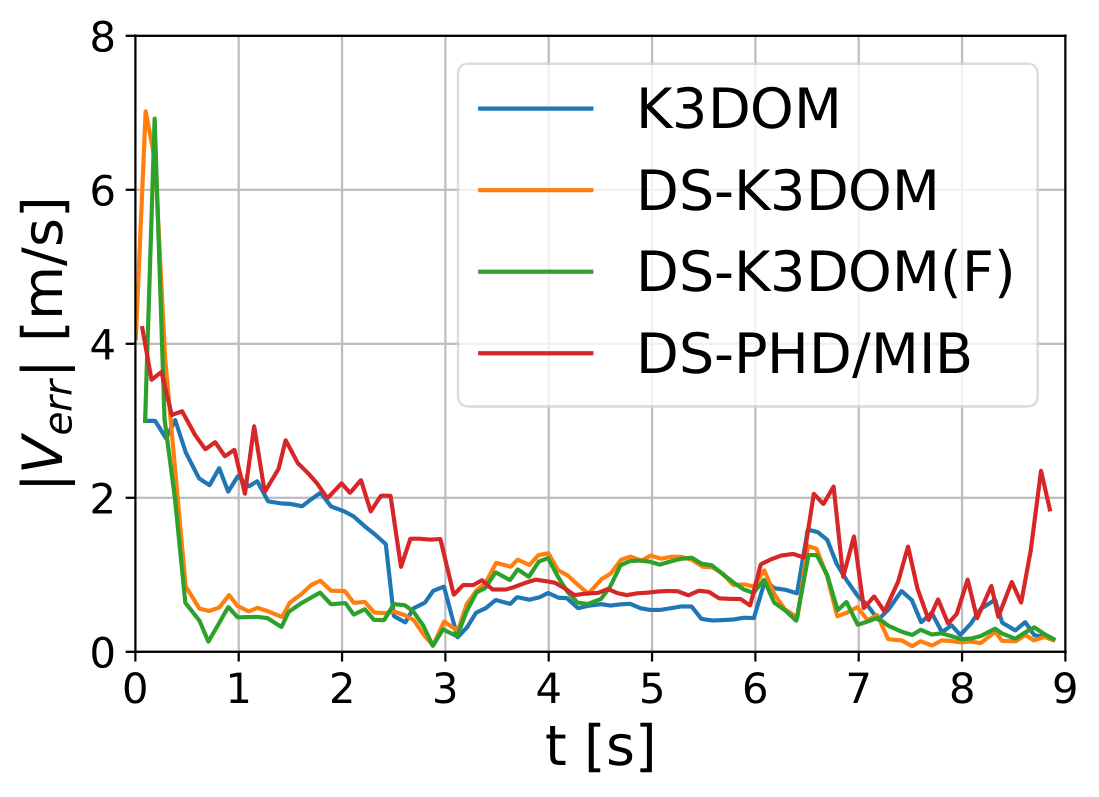}
      \caption{$\Vert V_{err} \Vert$ for Box 0}
      \label{fig:box0 vel}
    \end{subfigure}
    \hfill
    \begin{subfigure}{.19\textwidth}
      \centering
      \includegraphics[width=\linewidth]{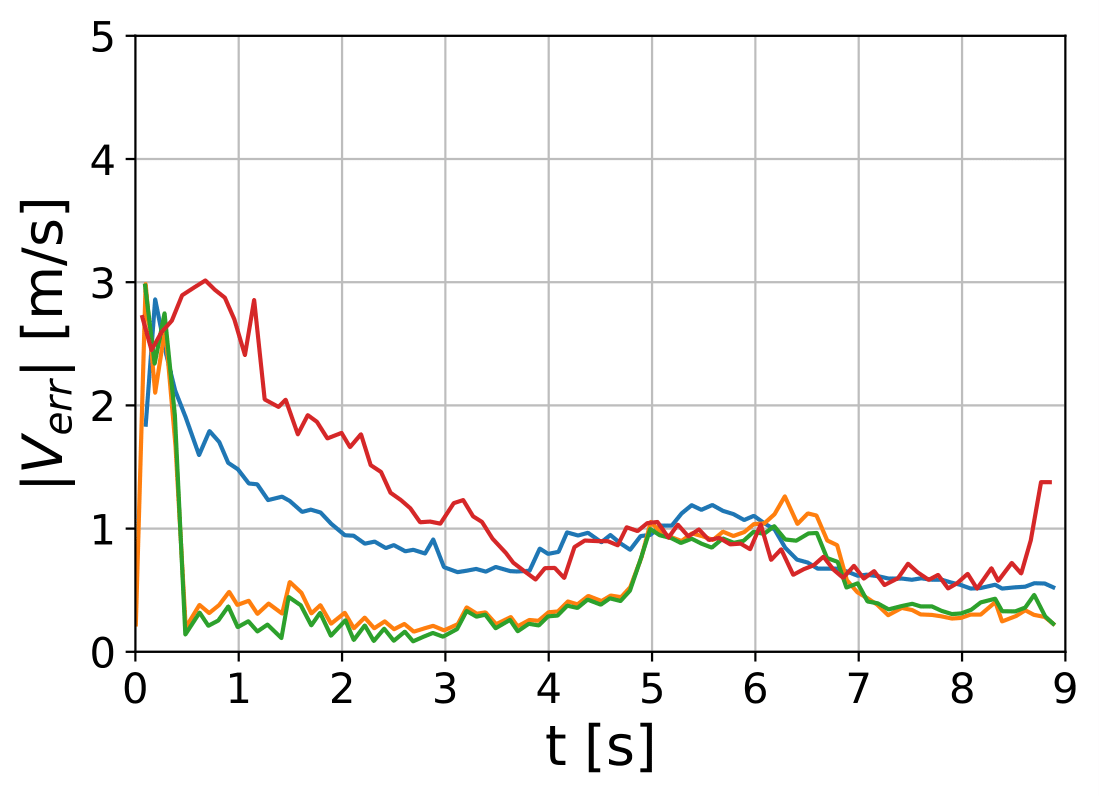}
      \caption{$\Vert V_{err} \Vert$ for Box 1}
      \label{fig:box1 vel}
    \end{subfigure}
    \hfill
    \begin{subfigure}{.19\textwidth}
      \centering
      \includegraphics[width=\linewidth]{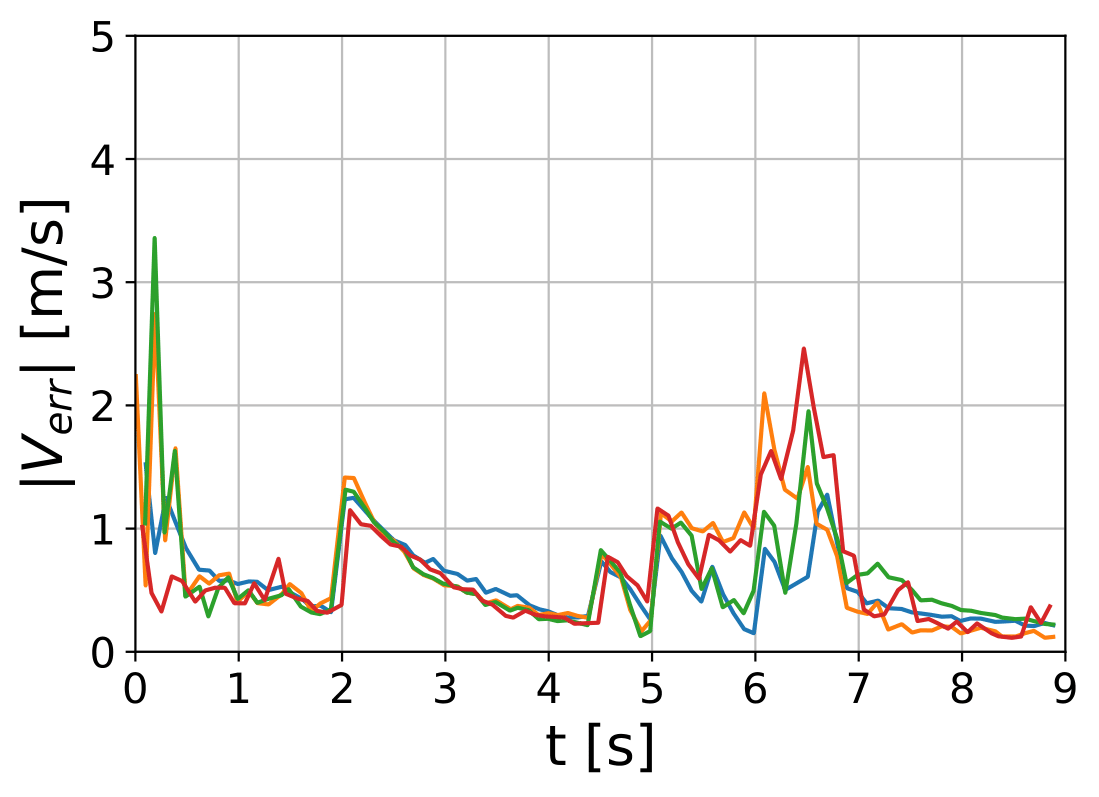}
      \caption{$\Vert V_{err} \Vert$ for Cylinder 0}
      \label{fig:cylinder0 vel}
    \end{subfigure}
\caption{Environment and results of simulation experiment. (a) shows \textit{Gazebo} simulation environment. (b-e) represent occupancy map of each algorithm. In (b-e), Colored dots indicate point clouds, Green cells represent `static (S)' state cells, and Blue color of cells represents `dynamic (D)' cells. (f-g) denote ROC curves of cell state classification (h-j) graph error magnitudes between the true velocity and the estimated velocity for each dynamic object.}
\label{fig:simulation}
\vspace{-.2in}
\end{figure*}

\section{Experiments}

\begin{table}[t]
    \centering
    \caption{Parameters for Experiments} \vspace{-.1in}
    \setlength{\tabcolsep}{0.3em}
    \begin{tabular}{|c|c|c|c|c|c|c|c|c|c|c|}
         \hline
         $\nu$ & $\nu_b$ & $\sigma_p$ & $\sigma_v$ & $l$ & $\sigma_0$ & $p_S$ & $p_B$ & $\gamma$ & $\beta$ & $R_0$ \\
         \hline
         $2 \!\times\! 10^6$ & $2 \!\times\! 10^5$ & 0.05m & 0.1m/s & 0.5m & 0.1 & 0.99 & 0.02 & 0.99 & 0.98 & 0.001\\
         \hline
    \end{tabular}
    \label{tab:parameters}
\vspace{-.2in}
\end{table}

\subsection{Setup}

The performance of DS-K3DOM is evaluated through simulations and real experiments in comparison to the baselines, the DS-PHD/MIB filter~\cite{Nuss2018} extended from the 2D to the 3D domain and K3DOM~\cite{Min2021}. The algorithms are implemented in \textit{ROS Melodic} using CUDA parallel computing and the simulations are conducted in \textit{Gazebo} with a desktop equipped with an Intel Core i7-8700 hexacore CPU and RTX 3060 Ti. The parameters shown in Table~\ref{tab:parameters} are used throughout the experiments, and the local map is set to span 40m x 40m x 5m for the simulations and 15m x 15m x 3m for the indoor experiments, with a resolution of 0.2m.

In simulations, a VLP-16 model of \textit{veolodyne\_simulator} package is adopted as a virtual LiDAR sensor model. The environment is designed with static and dynamic objects to evaluate the classification and velocity estimation performance of each algorithm. Static obstacles contain four brick buildings, while a LiDAR-eqquipped ego vehicle and dynamic objects `Box 0', `Box 1', and `Cylinder 0' are moving through an intersection as shown in Fig.~\ref{fig:simulation environment}.

The performances on classifying `occupied (O)' or `dynamic (D)' cells are assessed by evaluating cells that each algorithm decides to have sufficient observations on. The classification standards for the DS-PHD/MIB filter and K3DOM are described in \cite{Min2021}. For DS-K3DOM, it filters out cells with insufficient observations if $\zeta_0 < m^{(c)}(\Omega)$.
The threshold, $\zeta_0$, is empirically set to 0.5. Then, it classify each cell as
\begin{equation}
\label{occupied or dynamic}
\begin{cases}
    \text{`O'} & \text{if } P^{(c)}(D) + P^{(c)}(S) > \zeta_1\\
    \text{`D'} & \text{if } P^{(c)}(D) > \zeta_2 
\end{cases}
\end{equation}
for the pignistic probability $P^{(c)}(\cdot)$ in~\eqref{pignistic transformation}. The cell is correctly classified as `O' (resp. `D') if it contains any (resp. dynamic) objects. ROC (Receiver Operating Characteristic) curves and their AUC (Area Under Curve) are obtained by changing parameters $\zeta_1$ and $\zeta_2$.
The velocity of each dynamic object for DS-K3DOM is estimated in the same manner as \cite{Min2021}, by averaging velocities of all cells containing the object:
\begin{equation}
\label{objet velocity}
v^{obj} = 
    (\sum_{c \in obj}v^{(c)}\rho_{p}^{(c)})/(\sum_{c \in obj}\rho_{p}^{(c)}).
\end{equation}

In the indoor experiment, the performance of each algorithm is qualitatively assessed. The VLP-16 LiDAR is fixed by a support being apart from the ground, and a person walks around the LiDAR as shown in Fig. \ref{fig:indoor env}. A Jetson AGX Xavier is connected to the VLP-16 LiDAR to collect the point cloud data for the evaluation.

\subsection{Simulation}

Simulation results are shown in Fig.~\ref{fig:simulation}. To further improve the errors of misclassifying some ground cells as `D' due to sensor movement, we additionally test a variant of DS-K3DOM called `DS-K3DOM (F)'. It is implemented by removing particles on the ground assuming the prior knowledge on the ground position. The proposed algorithms are processed in about 4.5Hz speed which is real-time viable.

In Fig. \ref{fig:simulation dsphdmib} - \ref{fig:simulation ds_k3dom}, DS-K3DOM and DS-K3DOM (F) produce denser occupancy maps compared to the baselines. Specifically, the DS-PHD/MIB filter generates the sparsest occupancy map as shown in Fig. \ref{fig:simulation dsphdmib}. It also incorrectly classifies many static objects as `D', while the other algorithms do not. Meanwhile, K3DOM shows worse performance on classifying dynamic objects than the proposed methods. For example, in Fig.~\ref{fig:simulation k3dom}, `Box 1' and `Cylinder 0' are sparsely estimated, and, even worse, `Box 0' is invisible. Also, K3DOM produces some trail noise cells around `Box 1'. Unlike the baselines, the proposed algorithms densely estimate the dynamic objects as shown in Fig. \ref{fig:simulation naive_ds_k3dom} - \ref{fig:simulation ds_k3dom}.

However, in Fig. \ref{fig:occupy ROC}, all the baselines and the proposed algorithms show comparably high performances on the `O' classification. Their discrepancy with the qualitative results occurs as each method is assessed on a different set of cells determined by its own standard. For fair comparisons, we need to come up with a central standard to choose a common evaluation set. Similarly, Fig. \ref{fig:dynamic ROC} does not imply that the baselines have better dynamic object detection. Instead, the comparably high AUCs of the proposed methods with denser estimation (which implies larger evaluation sets) indicates better classification performances than the baselines.
In terms of the velocity estimation, as shown in Fig.~\ref{fig:box0 vel}-\ref{fig:cylinder0 vel}, the proposed algorithms reduce the errors faster and attain lower errors than the baselines.

\subsection{Indoor Experiment}

In Fig.~\ref{fig:indoor dsphdmib}-\ref{fig:indoor dsk3dom}, the estimation results on the person walking around the LiDAR are indicated in the red boxes. While the DS-PHD/MIB filter estimates the person as a static object with `S' cells, both K3DOM and DS-K3DOM recognize the person as a dynamic object. However, K3DOM produces trail noise for the person estimation unlike DS-K3DOM, and DS-K3DOM generates much denser occupancy map. 
Also, DS-K3DOM showed real-time capability, with a processing speed of around 7.5Hz on the desktop and 6Hz on the Jetson Xavier during the indoor experiment.

\section{Conclusion}
In the paper, a novel 3-D dynamic occupancy mapping algorithm, DS-K3DOM, was proposed by mathematically extending the PHD/MIB filter and approximating it in the DS domain with the kernel-based inference that enables dense occupancy mapping from sparse sensor measurements. Experiments have verified that DS-K3DOM shows outstanding performances with dense mapping and accurate velocity estimation compared to baselines, with the real-time capability. DS-K3DOM occasionally suffers from false static estimation on a large dynamic object. Future works may include heterogeneous sensor fusion or theoretical improvement for achieving a better performance.

\section*{Acknowledgment}
This research was supported by Unmanned Vehicles Core Technology Research and Development Program through the National Research  Foundation of Korea(NRF), Unmanned Vehicle Advanced Research Center(UVARC) funded by the Ministry of Science and ICT, the Republic of Korea (2020M3C1C1A01082375)





\bibliographystyle{unsrt}
\bibliography{reference}


\end{document}